\documentclass[runningheads,a4paper]{llncs}

\usepackage{amssymb}
\usepackage{graphicx}

\usepackage{url}
\urldef{\mailsa}\path|{chunweitian@163.com, yongxu@ymail.com, flksxm@126.com,yanke401@163.com}|
\newcommand{\keywords}[1]{\par\addvspace\baselineskip
\noindent\keywordname\enspace\ignorespaces#1}

\begin{document}
\bibliographystyle{plain}

\mainmatter  % start of an individual contribution

\title{Deep Learning for Image Denoising: A Survey}

\titlerunning{DLIDS: Chunwei Tian, Yong Xu, Lunke Fei, Ke Yan}

\author{Chunwei Tian \textsuperscript{1,2}
\and Yong Xu \textsuperscript{1,2,*}\and Lunke Fei\textsuperscript{3}\and Ke Yan\textsuperscript{1,2}\\}
\authorrunning{DLIDS: Chunwei Tian, Yong Xu, Lunke Fei, Ke Yan}

\institute{1. Bio-Computing Research Center, Harbin Institute of Technology, Shenzhen,\\
Shenzhen. 518055 Guangdong, China\\
2. Shenzhen Medical Biometrics Perception and Analysis Engineering Laboratory, Shenzhen, Shenzhen. 518055 Guangdong, China\\
3. School of Computers, Guangdong University of Technology, Guangzhou 510006, China\\
\mailsa\\
%\url{http://www.springer.com/lncs}}
}

\toctitle{Lecture Notes in Computer Science11}
\tocauthor{Authors' Instructions}
\maketitle

\begin{abstract}
Since the proposal of big data analysis and Graphic Processing Unit (GPU), the deep learning technology has received a great deal of attention and has been widely applied in the field of imaging processing. In this paper, we have an aim to completely review and summarize the deep learning technologies for image denoising proposed in recent years. Morever, we systematically analyze the conventional machine learning methods for image denoising. Finally, we point out some research directions for the deep learning technologies in image denoising.
\emph{abstract} environment.
\end{abstract}

\section{Introduction}

Image processing has numerous applications including image segmentation  \cite{shi2000normalized}, image classification \cite{qin2017weighted,wen2018low,tian2018fft,fei2018feature}, object detection \cite{girshick2014rich}, video tracking \cite{wang2015video}, image restoration \cite{zhang2017image} and action recognition \cite{wang2011action}. Especially, the image denoising technology is one of the most important branches of image processing technologies and is used as an ex-ample to show the development of the image processing technologies in last 20 years \cite{xu2015patch}. Buades et al. \cite{buades2005non} proposed a non-local algorithm method to deal with image denoising. Lan et al. \cite{lan2006efficient} fused the belief propagation inference method and Markov Random Fields (MRFs) to address image denoising. Dabov et al. \cite{dabov2007image} proposed to transform grouping similar two-dimensional image fragments into three-dimensional data arrays to improve sparisty for image denoising. These selection and extraction methods have amazing performance for image denoising. However, the conventional methods have two challenges \cite{zhang2017beyond}. First, these methods are non-convex, which need to manually set parameters. Second, these methods refer a complex optimization problem for the test stage, resulting in high computational cost.

In recent years, researches have shown that deep learning technologies can reply to deeper architecture to automatically learn and find more suitable image features rather than manual setting parameters, which effectively address drawbacks of traditional methods mentioned above \cite{krizhevsky2012imagenet}. Big data and GPU are also essential for deep learning technologies to improve the learning ability  \cite{hou2017deeply}. The learning ability of deep learning is finished by model (also referred to as network) and the model consists of many layers, including the convolutional layer, pooling layer, batch normalization layer and full connection layer. In other words, deep learning technologies can convert input data (e.g. images, speech and video) into outputs (e.g. object category, password unlocking and traffic information) by the model \cite{litjens2017survey}. Especially, convolutional neural network (CNN) is one of the most typical and successful deep learning network for image processing  \cite{lawrence1997face}. CNN was originated LeNet from 1998 and it was successfully used in hand-written digit recognition, achieving excellent performance \cite{lecun1989backpropagation}. However, convolutional neural networks (CNNs) haven¡¯t been widely used in other real applications before the arise of GPU and big data. In other words, the real success of CNNs attributed to ImageNet Large Scale Visual Recognition Challenge 2012 (ILSVRC 2012) where new CNN was proposed, named AlexNet and became a world champion in this ILSVRC 2012 \cite{krizhevsky2012imagenet,you2017100}.

In subsequent years, deeper neural networks have becoming popular and obtain promising performance for image processing \cite{simonyan2014very}. Karen Simonyan et al. \cite{simonyan2014very} in-creased the depth of neural networks to 16-19 weighted layers and convolution filter size of each layer was 3 $\times$ 3 for image recognition. Christian Szegedy et al. \cite{szegedy2015going} provided a mechanism by using sparsely connected layer  \cite{arora2014provable} instead of fully connected layers to increase the width and depth of the neural networks for image classification, named as Inception V1. Inception V1 effectively prevented to overfitting from enlarged size (width) of network and reduced the computing resource from increased depth of network. Previous researches show that the deep networks essentially use an end-to-end multilayer fashion to fuse different level fashion \cite{ioffe2015batch} and classifiers and the extracted features can be more robust by increasing the number of depth in networks. Despite deep networks have obtained successfully applications for image processing \cite{sermanet2013overfeat} , they can generate vanishing gradient or exploding gradient \cite{bengio1994learning} with increased network depth. That makes network hamper convergence. This problem can be solved by normalized initialization \cite{wu2018group}. However, when deeper neural networks get to converge, networks are saturated and degrade quickly with increasing depth of networks. The appearance of residual network effectively dealt with problems above for image recognition \cite{he2016deep}. ResNeXt method is tested to be very effectively for image classification \cite{xie2017aggregated}. Spatial-temporal Attention (SPA) method is very competitive for visual tracking \cite{zhu2017end}. Residual Dense Network (RDN) is also an effective tool for  image super-resolution \cite{zhang2018residual}. Furthermore, DiracNets \cite{zagoruyko2017diracnets}, IndRNN \cite{li2018independently} and varia-tional U-Net \cite{esser2018variational} also provide us with many competitive technologies for image pro-cessing. These deep networks are also widely applied in image denoising, which is the branch of image processing technologies. For example, the combination of kernel-prediction net and CNN is used to obtain denoised image \cite{bako2017kernel}. BMCNN utilizes NSS and CNN to deal with image denoising \cite{ahn2017block}. GAN is used to remove noise from noisy image \cite{tripathi2018correction}.

Although the researches above expose that deep learning technologies have ob-tained enormous success in the applications of image denoising, to own knowledge, there is no comparative study of deep learning technologies for image denoising. Deep learning technologies refer to properties of image denosing to propose wise solution methods, which are embedded in multiple hidden layers with end-end con-nection to better deal with them. Therefore, a survey is important and necessary to review the principles, performance, difference, merits, shortcomings and technical potential for image processing. Deeper CNNs (e.g. AlexNet, GoogLeNet, VGG and ResNet), which can show the ideas of deep learning technologies and successful rea-sons for image denoising. To better show the robustness of deep learning denoising, the performance of deep learning for image denoising is shown. The potential chal-lenges and directions of deep learning technologies for image denoising in the future are also offered in this paper.

The remainder of this paper is organized as follows. Section 2 overviews of typical deep learning methods. Section 3 provides deep learning technologies for image de-noising. Section 4 points out some potential research directions. Section 6 presents the conclusions of this paper.
\section{Typical deep network}
Nowadays, the most widely used model is trained with end-to-end in a supervised fashion, which is easy and simple to implement to train models. The popular network architecture is CNNs (ResNet). This network is widely used to deal with applications of image processing and obtain enormous success. The following sections will show the popular deep learning technology, discuss the merits and differences of the meth-od in Section 2.
\subsection{ResNet}
Deep CNNs have result in a lot of breakthroughs for image recognition. Especially, deep network plays an important role on image classification \cite{szegedy2015going}. Many other visual recognition applications are beneficial from deep networks. However, deeper network can have vanishing/exploding gradients \cite{szegedy2015going}. This problem has been effectively solved by normalized initialization \cite{tripathi2018correction}, which makes the network converge. When this network starts converging, performance of the network gets degraded. For exam-ple, the depth of this network are increased, the errors in the training model are in-creasing. The problem is effectively addressed by ResNet \cite{he2016deep}. The ideas of ResNet are that outputs of each two layers and their inputs are added as the new input. ResNet include many blocks and the block is shown in Fig.1, where $x$ and $f$ , respectively, denote input and activation function. A residual block is obtained by $f(x)$ + $x$ . The ResNet is popular based on the following reasons. First, ResNet is deep rather than width, which effectively controls the number of parameters and overcomes the overfitting problem. Second, it uses less pooling layers and more downsampling operations to improve transmission efficiency. Third, it uses BN and average pooling for regularization, which accelerate the speed of training model. Finally, it uses 3 $\times$ 3 filters of each convolutional layer to train model, which is faster than using the combination of 3 $\times$ 3 and 1 $\times$ 1 filters. As a result, ResNest takes the first place in ILSVRC 2015 and reduces 3.57$\%$ error on the ImageNet test set.

In addition, deformation networks of Residual network are popular and have been widely used in image classification, image denoising \cite{xie2018deep} and image resolution \cite{tai2017memnet}.
\begin{figure}
\centering
\includegraphics[height=4.2cm]{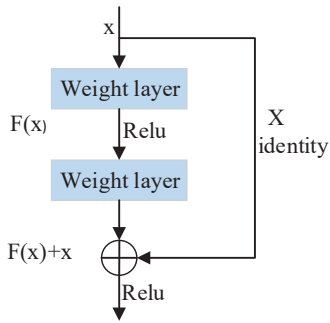}
\caption{Residual network: a building block}
\label{fig:example}
\end{figure}
\section{Image Denoising}
Image denoising is topic applications for image processing. We take image de-noising as an example to show the performance and principle for deep learning tech-nologies in image processing applications.

The aim of image denoising is to obtain clean image $x$ from a noisy image $y$ which is explained by $y = x + n$. $n$ denotes the additive white Gaussian noise (AWGN) with variance  $\sigma ^2$ . From the machine learning knowledge, we know that the image prior is an important for image denoising. In the past ten years, a lot of methods are proposed for model with image priors, such as Markov random filed (MRF) method \cite{lan2006efficient}, BM3D \cite{dabov2007image}, NCSR \cite{dong2013nonlocally} and NSS \cite{buades2008nonlocal}. Although these methods perform well for image denoising, they have two drawbacks. First, these methods need to optimize, which results in increasing computational cost. Second, these methods are non-convex, which need manual settings to improve performance. To address the problems, some discriminative learning schemes were proposed. A trainable nonlinear reaction diffusion method was proposed and used to learn image prior \cite{schmidt2014shrinkage}. A cascade of shrinkage fields fuse the random field-based model and half-quadratic algorithm into a single architecture \cite{zhang2018ffdnet}. Despite methods improve the performances for image denoising, they are limited to the specified forms of prior. Another shortcoming is that these methods can¡¯t use a model to deal with blind image denoising.

Deep learning technologies can effectively deal with problems above. And deep learning technologies are chosen for image denoising based on the following three-fold. First, they have deep architecture, which can learn more extractions. Second, BN and ReLu are added into deep architectures, which can accelerate the training speed. Third, networks of deep learning methods can run on GPU, which can train more samples and improve the efficiency. The proposed DnCNN \cite{zhang2017beyond} uses BN and ResNet to perform image denoising. This network not only deals with blind image denoising, but also addresses image super-resolution task and JPEG image deblocking. Its architecture is as shown in Fig. 2. Specifically, it obtains the residual image from the model and it needs to use $y = x + n$ to obtain clean image when it is in the test phase. It obtained PSNR of 29.13, which is higher than the state-of-the-art BM3D method of 28.57 for BSD68 dataset with $\sigma$  = 25.
\begin{figure}
\centering
\includegraphics[height=2.5cm]{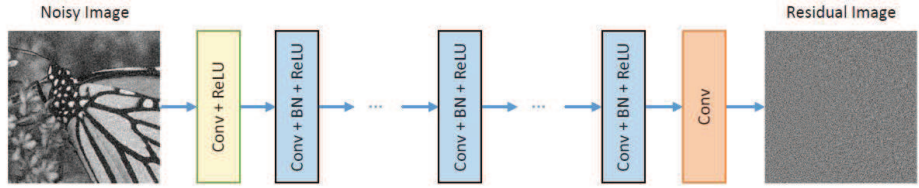}
\caption{The architecture of DnCNN}
\label{fig:example}
\end{figure}
\begin{figure}
\centering
\includegraphics[height=5cm]{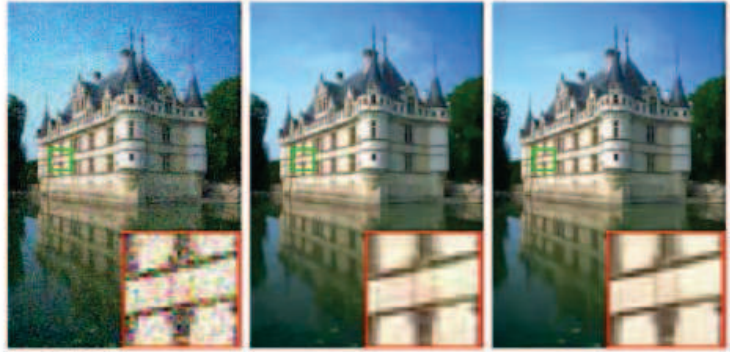}
\caption{Results of CBM3D and FFDNet for color image denoising (a)Noisy($\sigma$=35) (b)CBM3D(29.90dB) (c)FFDNet(30.51dB)}
\label{fig:example}
\end{figure}

FFDNet [46] uses noise level map and noisy image as input to deal with different noise levels. This method exploits a single model to deal with multiple noise levels. It is also faster than BM3D on GPU and CPU. As shown in Fig.3, performance of FFDNet outperforms the CBM3D \cite{lan2006efficient} method in image denoising. IRCNN \cite{zhang2018learning} fuses the model-based optimization method and CNN to address image denoising problem, which can deal with different inverse problems and multiple tasks with one single mode. In addition, it adds dilated convolution into network, which improves the per-formance for denoising. Its architecture is shown as Fig. 4.
\begin{figure}
\centering
\includegraphics[height=4cm]{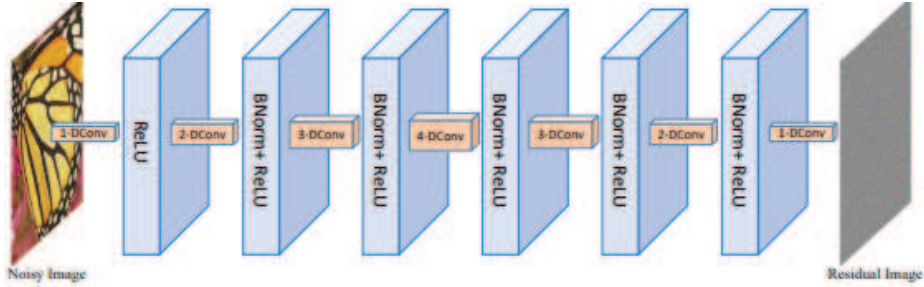}
\caption{The architecture of IRCNN}
\label{fig:example}
\end{figure}

In addition, many other methods also obtain well performance for image denoising. For example, fusion of the dilated convolution and ResNet is used for image denoising \cite{wang2017dilated}. It is a good choice for combing disparate sources of experts for image denosing \cite{choi2017integrating}. Universal denoising networks \cite{lefkimmiatis2018universal} for image denoising and deep CNN denoiser prior to eliminate multicative noise \cite{wang2017multiplicative} are also effective for image denoising. As shown in Table 1, deep learning methods are superior to the converntional methods. And the DnCNN method obtains excellent performance for image denoising.

\begin{table*}[!htb] \small
\caption{ Comparisons of different methods with $\sigma$  = 25  for image denoising.}
\label{tab:1}
\centering
\scalebox{1.5}[1.5]{
\begin{tabular}{|c|c|c|}
\hline
Methods  &PSNR & Dataset  \\
\hline
BM3D \cite{dabov2007image}	&28.57	&BSD68\\
\hline
WNNM \cite{gu2014weighted}	&28.83	&BSD68\\
\hline
TNRD \cite{chen2017trainable}	&28.92	&BSD68\\
\hline
DnCNN \cite{zhang2017beyond} &29.23	&BSD68\\
\hline
FFDNet \cite{zhang2018ffdnet} &29.19	&BSD68\\
\hline
IRCNN \cite{zhang2018learning} &29.15	&BSD68\\
\hline
DDRN \cite{wang2017dilated} &29.18	&BSD68\\
\hline
\end{tabular}}
\end{table*}

\section{Research directions}
\subsection{The challenges of deep learning technologies in image denoising}
According to existing researches, deep learning technologies achieve promising results in image denoising. However, these technologies also suffer from some challenges as follows.
(1)	Current deep learning denoising methods only deal with AWGN, which are not
effective for real noisy images, such as low light images.
(2)	They can¡¯t use a model to deal with all the low level vision tasks, such as image
denoising, image super-resolution, image blurring and image deblocking.
(3) They can¡¯t use a model to address the blind Gaussian noise.
\subsection{Some potential directions of deep learning technologies for image denoising}
According to the previous researches, deep learning technologies have the follow-ing changes for image denoising application above. First, deep learning technologies design different network architectures to deal with tasks above. Second, they can fuse the optimization and discrimination methods. Third, they can use multiple tasks to design the network. Fourth, they can change the input of the neural networks.
\section{Conclusion}
This paper first comprehensively introduces the development of deep learning technologies on image processing applications. And then shows the implementations of typical CNNs. After that, image denoising is illustrated in detail, which concludes the differences and ideas of different methods for image denoising in real world. Finally, this paper shows the challenges of deep learning methods for image processing applications and offers solutions. This review offers important cues on deep learning technologies for image processing applications. We believe that this paper could pro-vide researchers with a useful guideline working in the related fields, especially for the beginners worked in deep-learning.
\section{Acknowledgment}
This paper was supported in part by Shenzhen Municipal Science and Technology Innovation Council under Grant no. JCYJ20170811155725434, in part by the National Natural Science Foundation under Grant no. 61876051.
\bibliography{referencestcw}
%\end{spacing}
\end{document}